\documentclass{article}
\usepackage{arxiv}
\usepackage{amsmath}
\usepackage[utf8]{inputenc} 
\usepackage[T1]{fontenc}    
\usepackage{hyperref}       
\usepackage{url}            
\usepackage{booktabs}       
\usepackage{amsfonts}       
\usepackage{nicefrac}       
\usepackage{microtype}      
\usepackage{lipsum}		
\usepackage{graphicx}
\usepackage{natbib}
\usepackage{doi}
\usepackage{rotating}
\usepackage{multirow}

\title{ISMRNN: An Implicitly Segmented RNN Method with Mamba for Long-Term Time Series Forecasting}

\author{ {\includegraphics[scale=0.06]{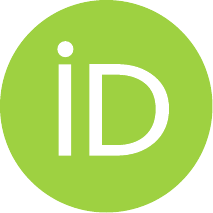}\hspace{1mm}Gaoxiang Zhao}\\
	Department of Mathematics\\
	Harbin Institue of Technology(Weihai)\\
	Weihai, Shandong\\
	\texttt{gaoxiang.zhao@stu.hit.edu.cn} \\
	\And
	{\includegraphics[scale=0.06]{orcid.pdf}\hspace{1mm}Li Zhou}\\
	Research Center for Data Hub and Security\\
	Zhejiang Lab\\
	Hangzhou, Zhejiang\\
	\texttt{zhou.li@zhejianglab.com} \\
	\And
	{\includegraphics[scale=0.06]{orcid.pdf}\hspace{1mm}Xiaoqiang Wang\textsuperscript{*}} \\
	Department of Mathematics and Statistics\\
	Shandong University\\
	Weihai, Shandong \\
\texttt{xiaoqiang.wang@sdu.edu.cn} \\
}

\renewcommand{\shorttitle}{}

\begin{document}
\maketitle

\begin{abstract}
Long time series forecasting aims to utilize historical information to forecast future states over extended horizons. Traditional RNN-based series forecasting methods struggle to effectively address long-term dependencies and gradient issues in long time series problems. Recently, SegRNN has emerged as a leading RNN-based model tailored for long-term series forecasting, demonstrating state-of-the-art performance while maintaining a streamlined architecture through innovative segmentation and parallel decoding techniques. Nevertheless, SegRNN has several limitations: its fixed segmentation disrupts data continuity and fails to effectively leverage information across different segments, the segmentation strategy employed by SegRNN does not fundamentally address the issue of information loss within the recurrent structure. To address these issues, we propose the ISMRNN method with three key enhancements: we introduce an implicit segmentation structure to decompose the time series and map it to segmented hidden states, resulting in denser information exchange during the segmentation phase. Additionally, we incorporate residual structures in the encoding layer to mitigate information loss within the recurrent structure. To extract information more effectively, we further integrate the Mamba architecture to enhance time series information extraction. Experiments on several real-world long time series forecasting datasets demonstrate that our model surpasses the performance of current state-of-the-art models.
\end{abstract}


\section{Introduction}
Long-term series forecasting is the process of predicting the future values of a sequence over an extended period. This task primarily aims to forecast long-term trends, which hold significant importance in domains such as climate prediction  \citep{bib1}, decision support \citep{bib2}, and policy formulation \citep{bib3}. Long-term sequence data often exhibit higher levels of uncertainty, which results in reduced prediction accuracy. Additionally, longer forecasting horizons require models to incorporate a broader historical context for accurate predictions, thereby increasing the complexity of modeling. Recent advancesments in long-term sequence analysis have shifted towards deep learning methods, highlighting the crucial need to develop diverse and effective time-series methodologies.

Long-term time series analysis involves identifying trends to understand past and predict future changes, managing uncertainty and noise to enhance prediction accuracy, and capturing the sequential relationships within the data. Commonly adopted feature representation methods in time series analysis include Convolutional Neural Networks(CNNs) \citep{bib4,bib5,bib6}, Multilayer Perceptrons(MLPs) \citep{bib8,bib9,bib10,bib26}, attention-based Transformer models \citep{bib11,bib12,bib22,bib23,bib24,bib27}, and Recurrent Neural Networks(RNNs) based on sequence structure \citep{bib13, bib25}. CNN-based methods use convolutional kernels to capture temporal and spatial sequences \citep{bib6}; however, the limited size of these kernels prevent the model from effectively capturing long-range dependencies in time series, thus limiting the expressiveness of the feature representation. Transformer-based methods, leveraging attention mechanisms to improve the capture of spatiotemporal sequences, thereby enhancing feature representation capabilities. MLPs have been exrensively applied in time series prediction, with models such as DLinear \citep{bib9}, which use simple linear layers and channel-independent methods, achieving superior performance compared to the advanced Transformer architectures of its time, prompting a reevaluation of the need for complex Transformer structures in time series prediction.

RNN-based methods have achieved remarkable success in long-term time series analysis tasks. The sequential structure of RNNs allows them to effectively handle short-term time series prediction tasks. However, as the length of the prediction sequence increases, traditional RNN structures gradually become ineffective due to their inability to handle long-term dependencies and gradient issues \citep{bib14}. To address this limitation, Shengsheng Lin et al. \citep{bib13} proposed the SegRNN model. By segmenting the time series and employing a parallel decoding strategy, SegRNN minimizes the number of RNN recurrent iterations while retaining sequential information to the greatest extent. This approach has led to significant success in both the accuracy and computational efficiency of long-term time series forecasting tasks. Nevertheless, SegRNN has several limitations: The segmentation process disrupts the continuity of the sequences, and information across different segments is not fully utilized. Additionally, the segmentation strategy effectively balances sequence information with iteration count; however, information loss within the recurrent structure increases as the prediction horizon extends, thereby limiting the model's capability.

In this work, we propose a novel model named ISMRNN to address the issues associated with SegRNN. Specifically, ISMRNN introduces an implicit segmentation structure that decomposes and maps the time series into encoded vectors through two linear transformations. This method facilitates more continuous processing during segmentation and enhances the utilization of information between different segments. Additionally, ISMRNN incorporates a residual structure with a linear layer, allowing some information to bypass the recurrent encoding structure, thus reducing information loss within the recurrent framework. Furthermore, we employ the Mamba structure\citep{bib17} for preprocessing the time series, which aids in capturing long-term dependencies more effectively. The main contributions of ISMRNN can be summarized as follows: 
\begin{itemize}
	\item
	Utilizing implicit segmentation for denser information exchange during the segmentation phase.
	\item
	Incorporating the Mamba structure to improve information preprocessing.
	\item
	The residual structure reduces information loss within the recurrent structure.
\end{itemize}

\section{Problem Statement}
The analysis of long-term time series involves addressing the challenge of predicting future values of multivariate time series based on historical data. Typically, the problem of time series analysis can be formulated as follows:

\begin{equation}
	\mathbf{Y} = F(\mathbf{X}) + \mathbf{\xi} ,
\end{equation}
in which $\mathbf{X} \in \mathbf{R}^{L \times C}$ represents multivariable time sequences with historical context, $F$ represents the trained neural network structure, $\mathbf{Y} \in \mathbf{R}^{H \times C}$ represents the future time sequences to be predicted, $L$ indicates the historical lookback window, $C$ indicates the number of variables, and $H$ indicates the time step to be predicted. Long-term time series analysis aims to expand the potential of predicting time steps, which poses significant challenges for training neural networks.

Long-term time series analysis and forecasting comprise two critical components: (1) Trend identification and analysis, which entail the recognition of long-term directional changes within the data. This component is crucial for understanding the behavior of time series, enabling the comprehension of past patterns and the prediction of potential future developments. (2) Uncertainty and noise management: Due to the extended time range of long-term time series data, they often exhibit increased levels of uncertainty and noise. Effectively managing and mitigating these uncertainties and noises is vital for improving the accuracy of predictions.

A significant amount of work has been dedicated to advancing the development of deep learning in time series analysis. These methods can be broadly divided into the following categories:

\subsection{Transformer Models}\label{subsec31}

The core of applying Transformer models to long-term time series tasks lies in their ability to capture long-term dependencies using the attention mechanism \citep{bib15}. The general form of the self-attention mechanism\citep{bib28} is given by:

\begin{equation}
	\text{Attention}(\mathbf{Q}, \mathbf{K}, \mathbf{V}) = \text{softmax}\left(\frac{\mathbf{Q} \mathbf{K}^T}{\sqrt{d_k}}\right) \mathbf{V}.
\end{equation}

Significant research efforts have advanced the application of Transformer architectures in long-term time series analysis. For instance, PatchTST\citep{bib11} introduces a method of segmenting time series into patches, reducing computational complexity and improving the model's ability to capture local semantic information. It also employs self-supervised learning, masking random patches and training the model to reconstruct them, enhancing forecasting accuracy. iTransformer\citep{bib16} addresses scalability by integrating hierarchical attention mechanisms and temporal convolutional networks, capturing both short-term and long-term dependencies more effectively. This lightweight architecture balances accuracy and efficiency, making it suitable for real-world applications. Crossformer \citep{bib12} employs a Dimension-segment-wise structure to embed time series into a two-dimensional vector, effectively capturing dependencies across time and dimensions through its Two-stage-attention structure. The segmentation technique utilized in Crossformer is particularly advantageous for long-term time series forecasting.

\subsection{MLP Models}\label{subsec32}

MLP methods have garnered significant attention in long-term time series forecasting due to their robust performance. Dlinear \citep{bib9} decomposes time series through the application of moving average kernels and remainder components, yielding two feature variables. These variables are then mapped to the prediction results via linear layers. The Dlinear model can be described as:

\begin{equation}
	\mathbf{\hat{Y}} = \mathbf{W}_1 (MA(\mathbf{X})) + \mathbf{W}_2 (R(\mathbf{X})),
\end{equation}
where $MA(\mathbf{X})$ represents the trend component obtained by the moving average kernel, $R(\mathbf{X})$ represents the remainder component, and $\mathbf{W}_1$, $\mathbf{W}_2$ are the weights of the linear layers.

TiDE \citep{bib10} introduces a residual structure within the linear layer to encode past time series and covariates, subsequently decoding the encoded time series with future covariates to obtain prediction results. The core operation of TiDE involves projecting features and encoding them through residual blocks, which can be described as:

\begin{equation}
	\mathbf{\hat{Y}} = \text{ResBlock}(\text{ReLU}(\mathbf{W_f X} + \mathbf{b_f}) + \text{Covariates}),
\end{equation}
where \(\mathbf{W_f}\) and \(\mathbf{b_f}\) denote the weights and biases of the projection layer, respectively. The \text{ResBlock} operation incorporates residual connections to preserve the integrity of the encoded features while integrating the covariates.

MLP-based methods for long-term time series analysis have demonstrated outstanding predictive accuracy while maintaining low time-space complexity, thus significantly promoting progress in the field of long-term time series forecasting.

\subsection{CNN Models}\label{subsec33}

The core of time series analysis based on CNN architectures lies in utilizing convolutional kernels to extract temporal and channel features. MICN \citep{bib4} aims to fully exploit the latent information in time series by designing a multi-scale hybrid decomposition module to decompose the complex patterns of input time series. Each module within this design leverages subsampling convolution and equidistant convolution to respectively extract local characteristics and global correlations. TimesNet \citep{bib5} transforms one-dimensional time series into a set of two-dimensional tensors based on multiple periodicities. It further introduces the TimesBlock to adaptively uncover these multi-periodic patterns and effectively extracts intricate temporal dynamics from the transformed tensors through parameter-efficient initializations. These methodologies have attained remarkable results in the domain of long-term time series forecasting.

\subsection{RNN Models}\label{subsec4}

The RNN structure is specifically designed for sequence tasks due to its ability to manage temporal dependencies through the recursive updating of hidden states. This mechanism enables the RNN to retain prior information and dynamically adjust internal states to capture short-term dependencies effectively. However, due to gradient issues, RNNs face challenges in capturing long-term dependencies. As the only RNN-based model specifically designed for long-term time series tasks, SegRNN has demonstrated exceptional performance. SegRNN addresses the challenge of retaining sequence information within the recurrent structure while minimizing information loss through two key methods:

\begin{itemize}
	\item SegRNN replaces point-wise iterations with segment-wise iterations. This approach balances the trade-off between retaining sequence information and the number of recurrent iterations, leading to improved performance.
	\item SegRNN substitutes recurrent multi-step forecasting with parallel multi-step forecasting. This parallel decoding approach enhances the model's predictive capability and inference speed.
\end{itemize}

By combining these two strategies, SegRNN achieves outstanding performance within a lightweight model structure and exhibits superior inference speed. However, the fixed segmentation approach of SegRNN limits the exchange of information between different segments and disrupts data continuity. Additionally, segmentation methods still face the challenge of information loss within the recurrent structure.

\section{Method}\label{sec4}
\subsection{Overview}\label{subsec41}

\begin{figure*}[ht]
	\centering
	\includegraphics[width=0.85\textwidth]{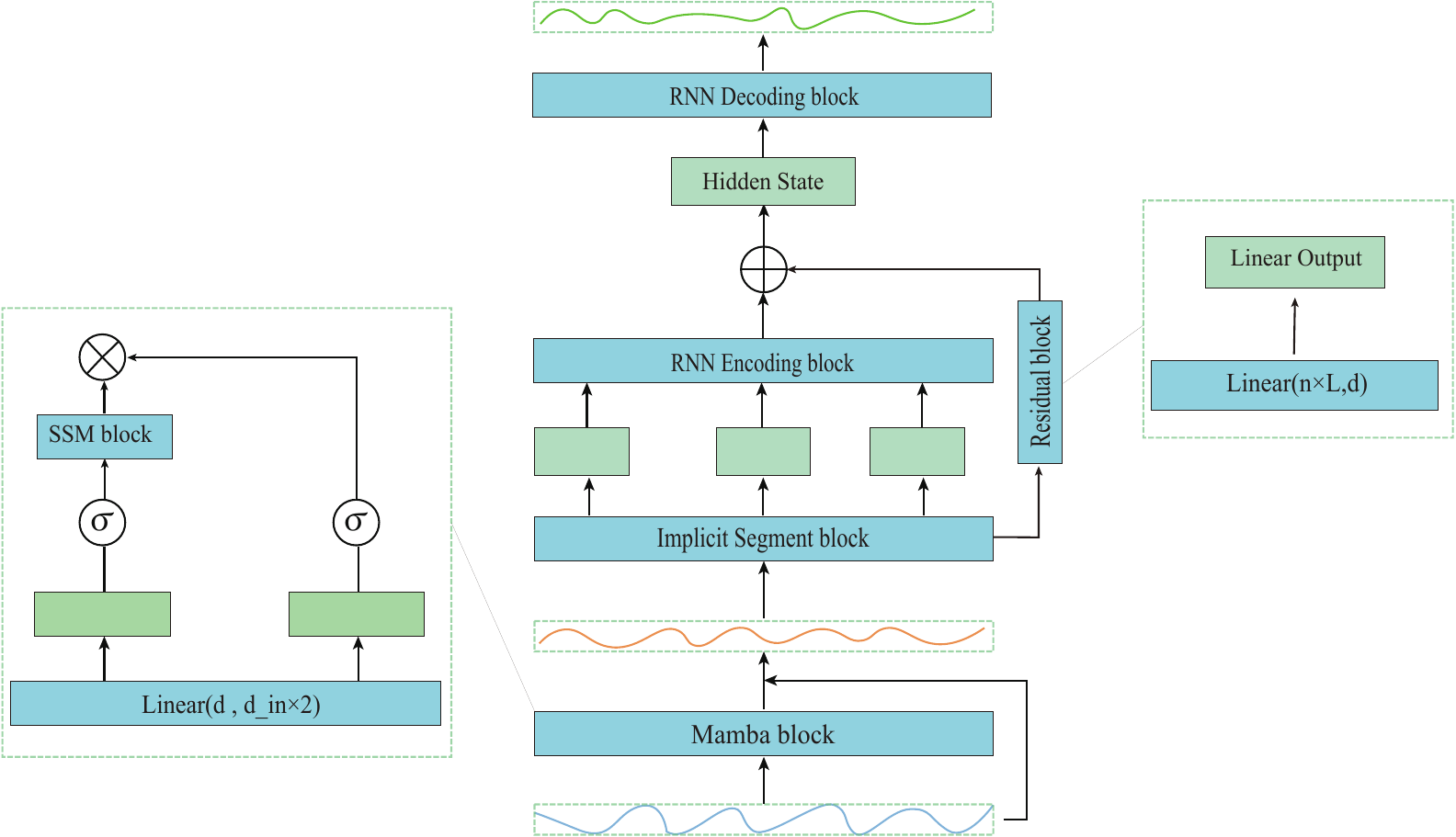}
	\caption{Overall structure of ISMRNN model.}
	\label{fig:Overall_model}
\end{figure*}

In this paper, we propose the ISMRNN method to overcome these limitations through three strategies, as illustrated in Figure \ref{fig:Overall_model}. We first introduce implicit segmentation, which allows for dense information exchange during the segmentation stage. Subsequently, we incorporate a residual structure to enable partial information bypassing the recurrent structure, thereby reducing information loss in recurrent structure. We also integrate the Mamba architecture to selectively process time series information, further enhancing the model's capability in handling temporal data.

\subsection{Implicit Segmentation}\label{subsec43}

The segmentation method employed by SegRNN is implemented by truncating the time dimension. For a given time series \( \mathbf{X}^{(i)} \in \mathbf{R}^{L \times C} \), selecting a segmentation length \( w \) results in segmented vectors of size  \(\mathbf{X}_w^{(i)} \in \mathbf{R}^{n \times w \times C} \), where \( n \) represents the number of segments.

\begin{equation}
	n = \frac{L}{w}.
\end{equation}

Subsequently, we use linear mapping to transform the dimension \( w \) into \( d \), obtaining the embedded hidden state of the segmented vectors \(\tilde{\mathbf{X}}_w^{(i)} \in \mathbf{R}^{n \times d \times C}
\). This segmentation method retains the sequence information while reducing the iteration count within the recurrent structure. However, this approach has some drawbacks. The simple truncation of data disrupts continuity, and there is no effective exchange of information between different segments, thus limiting the model's capability. 

\begin{figure*}[ht]
	\centering
	\includegraphics[width=0.9\textwidth]{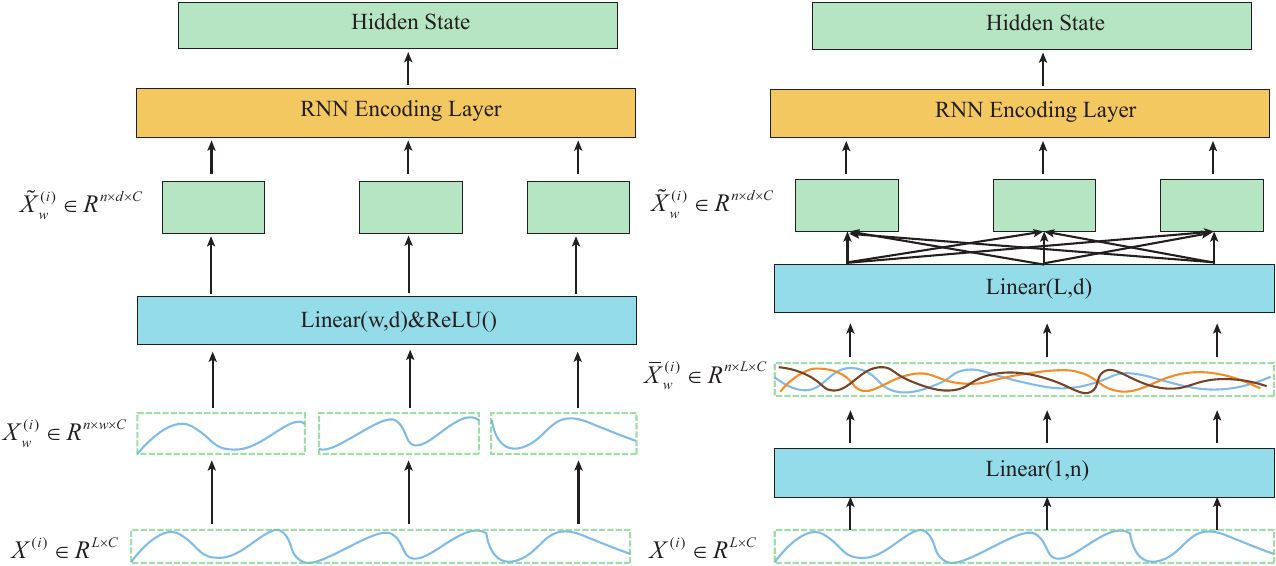}
	\caption{Comparsion of SegRNN Segmentation with Implicit Segmentation, we utilize a linear layers, from 1 to $n$, to decompose the time series. Subsequently, a second mapping is applied to compress the dimensions. Note that each segmented vector embedding can encompass information from other segments of the original time series.
	}
	\label{fig:Implicit Segmentation}
\end{figure*}

To address these issues, we propose an implicit segmentation strategy, as illustrated in Figure \ref{fig:Implicit Segmentation}. This strategy initially decomposes the time series into \(\bar{\mathbf{X}}^{(i)} \in \mathbf{R}^{n \times L \times C}\) through linear mapping and subsequently transforms the information into the embedded hidden state \(\tilde{\mathbf{X}}_w^{(i)} \in \mathbf{R}^{n \times d \times C}\). Notably, after segmentation, each segment includes both the information of that segment and some additional information, ensuring more continuous information processing and thereby enhancing the model's performance.

Furthermore, our proposed method does not explicitly define the segmentation length. Instead, it achieves data segmentation and dimensionality transformation through dual linear projections. The incorporation of linear fully connected layers facilitates a more continuous and dense transformation of information. It is importanct to emphasize that this implicit segmentation strategy is not confined to a singe model but rather represents a general and straightforward technique for information enhancement.

\subsection{Residual Structure}\label{subsec44}
Extensive experiments have demonstrated that increasing the number of iterations within the recurrent structure can negatively impact the model’s information retention. The use of segmentation techniques can reduce the number of iterations in the recurrent structure from \( L \) to \( n \).
Specifically, for the vector obtained through segmentation in SegRNN, \(\tilde{\mathbf{X}}_w^{(i)}  \in \mathbf{R}^{n \times d \times C}\), we transform it in the RNN structure with a hidden layer dimension \(d\) using the following formula:

\begin{equation}
	\begin{aligned}
		\mathbf{h}_0 &= \mathbf{0}, \\
		\mathbf{h}_1 &= f(\mathbf{h}_0,\tilde{\mathbf{X}}_w^{(i)}[1, :, :]; \mathbf{W}), \\
		\mathbf{h}_2 &= f(\mathbf{h}_1, \tilde{\mathbf{X}}_w^{(i)}[2, :, :]; \mathbf{W}), \\
		&\vdots \\
		\mathbf{h}_n &= f(\mathbf{h}_{n-1}, \tilde{\mathbf{X}}_w^{(i)}[n, :, :]; \mathbf{W}),
	\end{aligned}
\end{equation}
where \(\mathbf{h}_t \in \mathbf{R}^{1 \times C \times d}\) represents the hidden state at step \(t\), \(\tilde{\mathbf{X}}_w^{(i)}[t, :, :] \) represents the input vector at step \(t\), \(\mathbf{W}\) denotes the RNN's weight parameters, and \(f\) is the RNN's activation function.

However, relying solely on segmentation does not fundamentally address the issue of information loss within the recurrent structure, which may become more pronounced as the prediction length increases. To address this issue, we incorporate a residual structure into the model. For the vector \(\bar{\mathbf{X}}^{(i)} \in \mathbf{R}^{n \times L \times C}\) obtained through the decomposition by the first linear layer , we apply a linear mapping to \(\bar{\mathbf{X}}^{(i)}\), resulting in \(\mathbf{h}'_n\). Finally, we add \(\mathbf{h}'_n\) to the corresponding position of the RNN encoder's output vector to obtain the final encoder output vector \(\mathbf{h}_n\).

Introducing the residual structure allows some information to skip the recurrent structure and be directly mapped to the encoder output. This approach minimizes information loss within the recurrent structure, providing a beneficial enhancement for time series data.

\subsection{Mamba structure}\label{subsec42}

The Mamba architecture excels in various fields, including large language models and long-term time series forecasting, due to its selective state space models (SSMs) that enable content-based parameter adjustment. This feature allows the architecture to dynamically select and process relevant information, making it highly effective for distinguishing signals from noise, particularly in time series forecasting.

We incorporate the Mamba architecture at the beginning of our workflow to preprocess the time series data. Specifically, we use the SSM component, which adjusts parameters based on the input signal. This setup enhances the model’s ability to selectively propagate and forget information along the sequence, focusing on relevant data points to improve forecasting accuracy.

Notably, we deviate from the standard Mamba architecture by excluding the convolutional layer, except for weather datasets. This choice is based on performance considerations, as we believe segmentation techniques effectively capture local information without the added computational load of convolutional processing.

\section{Experiments}\label{sec5}
Our experiments are structured as follows. We first introduce the datasets used for long-term time series analysis. Subsequently, we provide a detailed description of the experimental setup and baseline models for comparison. Then we presents and analyzes the performance metrics. We further conducted ablation studies to investigate the effectiveness of the model. Finally, we evaluated the model's efficiency to demonstrate its high spatiotemporal efficiency.
All experiments in this section were conducted on two NVIDIA T4 GPU.

\subsection{Datasets}\label{subsec51}
The evaluation is conducted across six real-world datasets spanning various domains, including power transformer temperature, weather, electricity load, with channel ranging from 7 to 321 and frequencies from every 10 minutes to hourly recordings. Detailed dataset information is provided in Table \ref{tab:datasets}.

\begin{table}[htbp]
	\centering
	\caption{Dataset Basic Attributes}
	\label{tab:datasets}
	\renewcommand{\arraystretch}{1} 
	\fontsize{9}{11}\selectfont 
	\begin{tabular}{llccccc}
		\hline
		Data & Electricity & Weather & ETTh1 & ETTh2 & ETTm1 & ETTm2\\
		\hline
		Chan. & 321 & 21 & 7 & 7 & 7 & 7\\
		\hline
		Freq. & 1 hour  & 10 mins & 1 hour & 1 hour & 15 mins & 15 mins\\
		\hline
		Points & 26,304 & 52,696 & 17,420 & 17,420 & 69,680 & 69,680\\
		\hline
	\end{tabular}
	\setlength{\tabcolsep}{6pt} 
\end{table}

\subsection{Experimental setup and Baselines}\label{subsec52}

The unified configuration of the model is substantially aligned with the SegRNN approach. The look-back window is set to be 96, a single layer of Mamba is utilized for preprocessing the data and a single GRU layer is used for sequence processing. The dimensionality of hidden layer with GPU structure is set to be 512, and the training epochs are set to be 30. Dropout rate, learning rate, and batch size vary with the data and the scale of the model.

As baselines, we have selected state-of-the-art and representative models in the long-term time series forecasting domain, comprising the following categories: (i) RNNs: SegRNN; (ii) MLPs: Dlinear; (iii) CNNs: TimesNet; (iv) Transformers: iTransformer, Crossformer, PatchTST.

\subsection{Main Result}\label{subsec53}

We select two evaluation metrics, MSE and MAE, with prediction steps of 96, 192, 336, and 720 time steps. The resulting forecasts are presented in Table \ref{tab:comparison}. Note that all data, except for our method, originate from 
official code repositories and original papers.

Our method achieved top positions in 36 out of 48 metrics, while the SegRNN method and the iTransformer, each achieved top position in 7 and 6 metrics, respectively. This demonstrates the powerful capabilities of our model.

\begin{table*}[ht]
	\centering
	\caption{Performance metrics for Long-time series analysis.}
	\label{tab:comparison}
	\setlength{\tabcolsep}{1.78mm}
	\fontsize{8}{10}\selectfont 
	\renewcommand{\arraystretch}{1.35} 
	
	\begin{tabular}{c|c|cc|cc|cc|cc|cc|cc|cc}
		\hline
		\multicolumn{1}{c|}{Datasets} &
		\multicolumn{1}{c|}{} & \multicolumn{4}{c|}{RNNs} & \multicolumn{6}{c|}{Transformers} & \multicolumn{2}{c|}{MLPs} & \multicolumn{2}{c}{CNNs} \\
		\hline
		& & \multicolumn{2}{c|}{ISMRNN} & \multicolumn{2}{c|}{SegRNN} & \multicolumn{2}{c|}{iTransformer} & \multicolumn{2}{c|}{Crossformer} & \multicolumn{2}{c|}{PatchTST} & \multicolumn{2}{c|}{Dlinear} & \multicolumn{2}{c}{TimesNet} \\
		\hline
		& & MSE & MAE & MSE & MAE & MSE & MAE & MSE & MAE & MSE & MAE & MSE & MAE & MSE & MAE  \\
		\hline
		& 96 & \textbf{0.365} & \textbf{0.384} & 0.368 & 0.395 & 0.386 & 0.405 & 0.423 & 0.448 & 0.414 & 0.419 & 0.386 & 0.400 & 0.384 & 0.402  \\
		ETTh1 & 192 & 0.415 & \textbf{0.413} & \textbf{0.408} & 0.419 & 0.441 & 0.436 & 0.471 & 0.474 & 0.460 & 0.445 & 0.437 & 0.432 & 0.436 & 0.429  \\
		& 336 & 0.463 & 0.441 & \textbf{0.444} & \textbf{0.440} & 0.487 & 0.458 & 0.570 & 0.546 & 0.501 & 0.466 & 0.481 & 0.459 & 0.491 & 0.469 \\
		& 720 & 0.468 & 0.460 & \textbf{0.446} & \textbf{0.457} & 0.503 & 0.491 & 0.653 & 0.621 & 0.500 & 0.488 & 0.519 & 0.516 & 0.521 & 0.500  \\
		\hline
		& 96 & \textbf{0.275} & \textbf{0.326} & 0.278 & 0.335 & 0.297 & 0.349 & 0.745 & 0.584 & 0.302 & 0.348 & 0.333 & 0.387 & 0.340 & 0.374 \\
		ETTh2 & 192 & \textbf{0.354} & \textbf{0.378} & 0.359 & 0.389 & 0.380 & 0.400 & 0.877 & 0.656 & 0.388 & 0.400 & 0.477 & 0.476 & 0.402 & 0.414 \\
		& 336 & \textbf{0.404} & \textbf{0.416} & 0.421 & 0.436 & 0.428 & 0.432 & 1.043 & 0.731 & 0.426 & 0.433 & 0.594 & 0.541 & 0.452 & 0.452  \\
		& 720 & \textbf{0.407} & \textbf{0.428} & 0.432 & 0.455 & 0.427 & 0.445 & 1.104 & 0.763 & 0.431 & 0.446 & 0.831 & 0.657 & 0.462 & 0.468 \\
		\hline
		& 96 & \textbf{0.314} & \textbf{0.345} & 0.330 & 0.369 & 0.334 & 0.368 & 0.404 & 0.426 & 0.329 & 0.367 & 0.345 & 0.372 & 0.338 & 0.375  \\
		ETTm1 & 192 & \textbf{0.361} & \textbf{0.374} & 0.369 & 0.392 & 0.377 & 0.391 & 0.450 & 0.451 & 0.367 & 0.385 & 0.380 & 0.389 & 0.374 & 0.387  \\
		& 336 & \textbf{0.388} & \textbf{0.392} & 0.399 & 0.412 & 0.426 & 0.420 & 0.532 & 0.515 & 0.399 & 0.410 & 0.413 & 0.413 & 0.410 & 0.411 \\
		& 720 & \textbf{0.448} & \textbf{0.430} & 0.454 & 0.443 & 0.491 & 0.459 & 0.666 & 0.589 & 0.454 & 0.439 & 0.474 & 0.453 & 0.478 & 0.450\\
		\hline
		& 96 & \textbf{0.170} & \textbf{0.248} & 0.173 & 0.255 & 0.180 & 0.264 & 0.287 & 0.366 & 0.175 & 0.259 & 0.193 & 0.292 & 0.187 & 0.267 \\
		ETTm2 & 192 & \textbf{0.233} & \textbf{0.292} & 0.237 & 0.298 & 0.250 & 0.309 & 0.414 & 0.492 & 0.241 & 0.302 & 0.284 & 0.362 & 0.249 & 0.309  \\
		& 336 & \textbf{0.294} & \textbf{0.331} & 0.296 & 0.336 & 0.311 & 0.348 & 0.597 & 0.542 & 0.305 & 0.343 & 0.369 & 0.427 & 0.321 & 0.351  \\
		& 720 & 0.391 & \textbf{0.389} & \textbf{0.389} & 0.407 & 0.412 & 0.407 & 1.024 & 0.679 & 0.394 & 0.390 & 0.387 & 0.370 & 0.400 & 0.388 \\
		\hline
		& 96 & 0.149 & 0.242 & 0.151 & 0.245 & \textbf{0.148} & \textbf{0.240} & 0.219 & 0.314 & 0.181 & 0.270 & 0.197 & 0.282 & 0.168 & 0.272 \\
		Electricity & 192 & 0.165 & 0.259 & 0.164 & 0.258 & \textbf{0.162} & \textbf{0.253} & 0.231 & 0.322 & 0.188 & 0.274 & 0.196 & 0.285 & 0.184 & 0.289 \\
		& 336 & 0.181 & 0.275 & 0.180 & 0.277 & \textbf{0.178} & \textbf{0.269} & 0.246 & 0.337 & 0.204 & 0.293 & 0.209 & 0.301 & 0.198 & 0.300 \\
		& 720 & \textbf{0.218} & \textbf{0.307} & \textbf{0.218} & 0.313 & 0.225 & 0.317 & 0.280 & 0.363 & 0.246 & 0.324 & 0.245 & 0.333 & 0.220 & 0.320 \\
		\hline
		& 96 & \textbf{0.147} & \textbf{0.187} & 0.165 & 0.227 & 0.174 & 0.214 & 0.158 & 0.230 & 0.177 & 0.210 & 0.196 & 0.255 & 0.172 & 0.220  \\
		Weather & 192 & \textbf{0.198} & \textbf{0.238} & 0.211 & 0.273 & 0.221 & 0.254 & 0.206 & 0.277 & 0.225 & 0.250 & 0.237 & 0.296 & 0.219 & 0.261  \\
		& 336 & \textbf{0.256} & \textbf{0.282} & 0.270 & 0.318 & 0.278 & 0.296 & 0.272 & 0.335 & 0.278 & 0.290 & 0.283 & 0.335 & 0.280 & 0.306  \\
		& 720 & \textbf{0.334} & \textbf{0.337} & 0.357 & 0.376 & 0.358 & 0.349 & 0.398 & 0.418 & 0.354 & 0.340 & 0.345 & 0.381 & 0.365 & 0.359 \\
		\hline
		\multicolumn{2}{r|}{Count} & \multicolumn{2}{c|}{36} & \multicolumn{2}{c|}{7} & \multicolumn{2}{c|}{6} & \multicolumn{2}{c|}{0} & \multicolumn{2}{c|}{0} & \multicolumn{2}{c|}{0} & \multicolumn{2}{c}{0} \\
		\hline
	\end{tabular}
\end{table*}

We subsequently conducted experiments to observe how the model's capability varies with changes in the look-back window. The hyperparameter settings of the method were kept consistent with look-back window 96, and the horizon was uniformly set to 336. The evaluation was performed on the ETTh2 dataset, with the MSE and MAE losses of the method shown in the Figure \ref{fig:SegRNN_vs_ISMRNN}:

\begin{figure}[ht]
	\centering
	\includegraphics[width=0.8\textwidth]{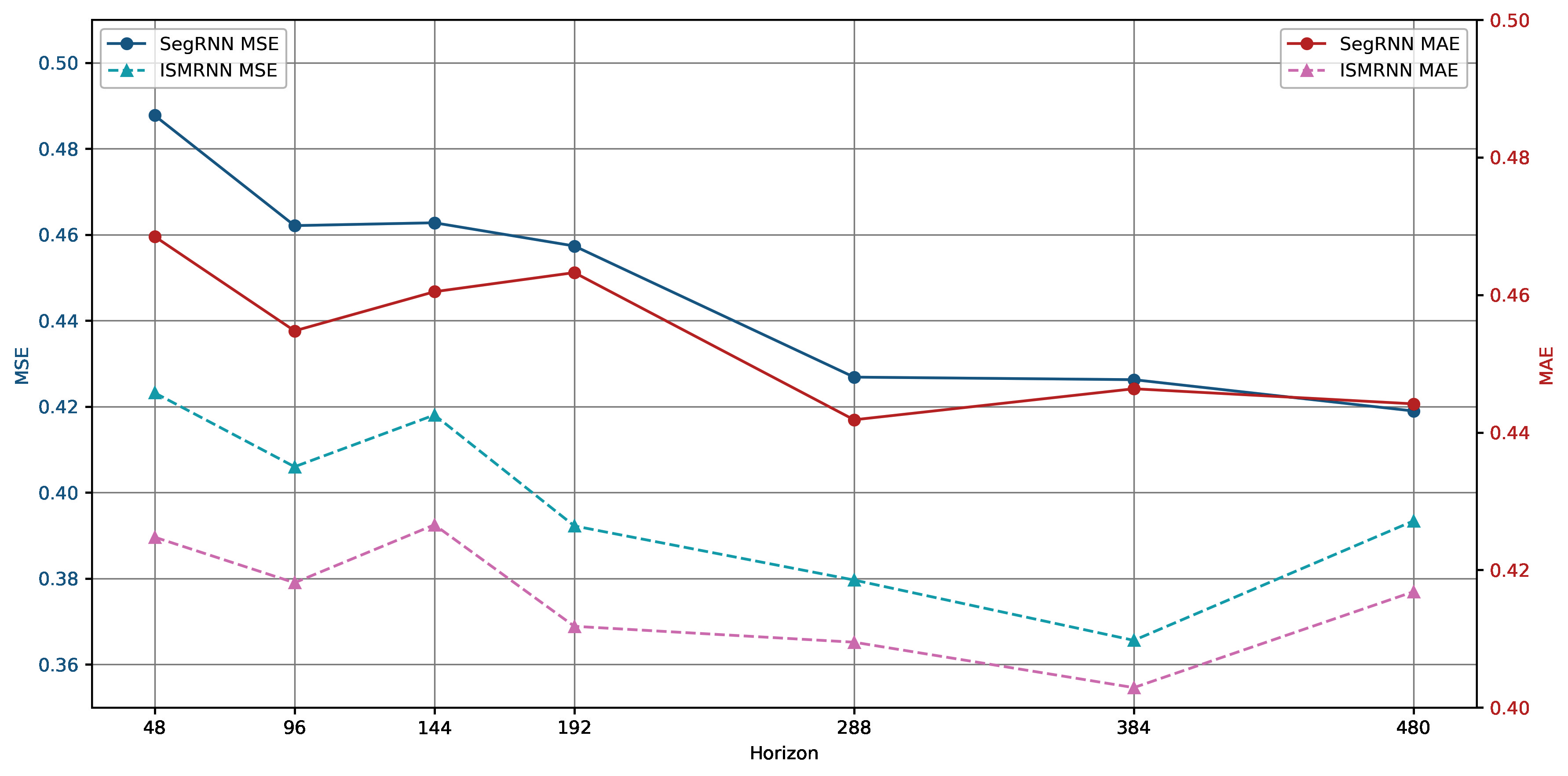}
	\caption{Comparsion of SegRNN and ISMRNN with vary look-back window length.}
	\label{fig:SegRNN_vs_ISMRNN}
\end{figure}

Our model exhibited smaller losses across various look-back windows, demonstrating the superiority of our method. Notably, the advantage of our method is more pronounced with smaller look-back windows. As the look-back window increases, this advantage diminishes, possibly due to noise introduced by segmentation affecting the model. Nevertheless, our model maintained a considerable advantage.

\subsection{Ablation study}\label{subsec54}

We further conduct an ablation study to verify the effectiveness of incorporating the Mamba structure, implicit segmentation, and residual structures into our model. We performed experiments excluding the Mamba structure and implicit segmentation structure respectivly. For clarity, in the table below, "M" denotes preprocessing using the Mamba structure, and "LR" represents the addition of implicit segmentation and residual structure. The results are shown in Table \ref{table:results}.

\begin{table*}[ht]
	\centering
	\caption{Ablation study on different structure}
	\label{table:results}
	\fontsize{9}{11}\selectfont 
	\renewcommand{\arraystretch}{1.35} 
	
	\begin{tabular}{c|c|cc|cc|cc|cc}
		\hline
		Datasets & Models & \multicolumn{2}{c|}{M\&LR\&SegRNN} & \multicolumn{2}{c|}{LR\&SegRNN} & \multicolumn{2}{c|}{M\&SegRNN} & \multicolumn{2}{c}{SegRNN} \\
		& & MSE & MAE & MSE & MAE & MSE & MAE & MSE & MAE \\
		\hline
		& 96 & \textbf{0.147} & \textbf{0.187} & 0.163 & 0.202 & 0.191 & 0.232 & 0.165 & 0.227 \\
		Weather & 192 & \textbf{0.199} & \textbf{0.238} & 0.207 & 0.242 & 0.243 & 0.277 & 0.211 & 0.273 \\
		& 336 & \textbf{0.256} & 0.282 & 0.264 & \textbf{0.287} & 0.305 & 0.324 & 0.270 & 0.318 \\
		& 720 & \textbf{0.332} & \textbf{0.337} & 0.350 & 0.342 & 0.380 & 0.377 & 0.357 & 0.376 \\
		\hline
		& 96 & \textbf{0.365} & \textbf{0.384} & 0.377 & 0.387 & 0.480 & 0.446 & 0.368 & 0.395 \\
		ETTh1 & 192 & 0.415 & \textbf{0.413} & 0.423 & 0.416 & 0.520 & 0.472 & \textbf{0.408} & 0.419 \\
		& 336 & 0.463 & 0.441 & 0.475 & 0.437 & 0.543 & 0.491 & \textbf{0.444} & \textbf{0.440} \\
		& 720 & 0.468 & 0.460 & 0.476 & 0.454 & 0.549 & 0.515 & \textbf{0.446} & \textbf{0.457} \\
		\hline
		& 96 & \textbf{0.275} & \textbf{0.326} & 0.281 & 0.327 & 0.326 & 0.368 & 0.278 & 0.335 \\
		ETTh2 & 192 & \textbf{0.354} & \textbf{0.378} & 0.368 & 0.385 & 0.418 & 0.420 & 0.359 & 0.389 \\
		& 336 & 0.404 & \textbf{0.416} & \textbf{0.402} & \textbf{0.416} & 0.462 & 0.456 & 0.421 & 0.436 \\
		& 720 & \textbf{0.407} & \textbf{0.428} & 0.421 & 0.435 & 0.467 & 0.472 & 0.432 & 0.455 \\
		\hline
		& 96 & \textbf{0.314} & \textbf{0.345} & 0.315 & 0.346 & 0.535 & 0.454 & 0.330 & 0.369 \\
		ETTm1 & 192 & \textbf{0.361} & \textbf{0.374} & 0.362 & 0.376 & 0.556 & 0.469 & 0.369 & 0.392 \\
		& 336 & \textbf{0.388} & \textbf{0.392} & 0.390 & 0.395 & 0.576 & 0.480 & 0.399 & 0.412 \\
		& 720 & 0.448 & 0.430 & \textbf{0.446} & \textbf{0.427} & 0.631 & 0.509 & 0.454 & 0.443 \\
		\hline
		& 96 & \textbf{0.170} & \textbf{0.248} & 0.173 & 0.250 & 0.215 & 0.298 & 0.173 & 0.255 \\
		ETTm2 & 192 & \textbf{0.233} & \textbf{0.292} & 0.235 & 0.293 & 0.283 & 0.339 & 0.237 & 0.298 \\
		& 336 & \textbf{0.294} & \textbf{0.331} & 0.297 & 0.332 & 0.352 & 0.381 & 0.296 & 0.336 \\
		& 720 & 0.391 & 0.389 & 0.392 & \textbf{0.388} & 0.463 & 0.441 & \textbf{0.389} & 0.407 \\
		\hline
		\multicolumn{2}{c|}{Average} & \textbf{0.334} & \textbf{0.355} & 0.341 & 0.357 & 0.425 & 0.411 & 0.340 & 0.372 \\
		\hline
	\end{tabular}
\end{table*}

The ablation study reveals that the model incorporating the Mamba structure, along with implicit segmentation and
residual structures performs the best overall. This is followed by the model that includes only implicit segmentation and
residual structures. Interestingly, merely adding the Mamba structure does not significantly enhance model performance.
This may be due to a synergistic effect between the two structures. Consequently, the ablation study confirms the
superiority of our proposed structure.

\subsection{Model efficiency}\label{subsec55}
\begin{figure}[ht]
	\centering
	\includegraphics[width=0.8\textwidth]{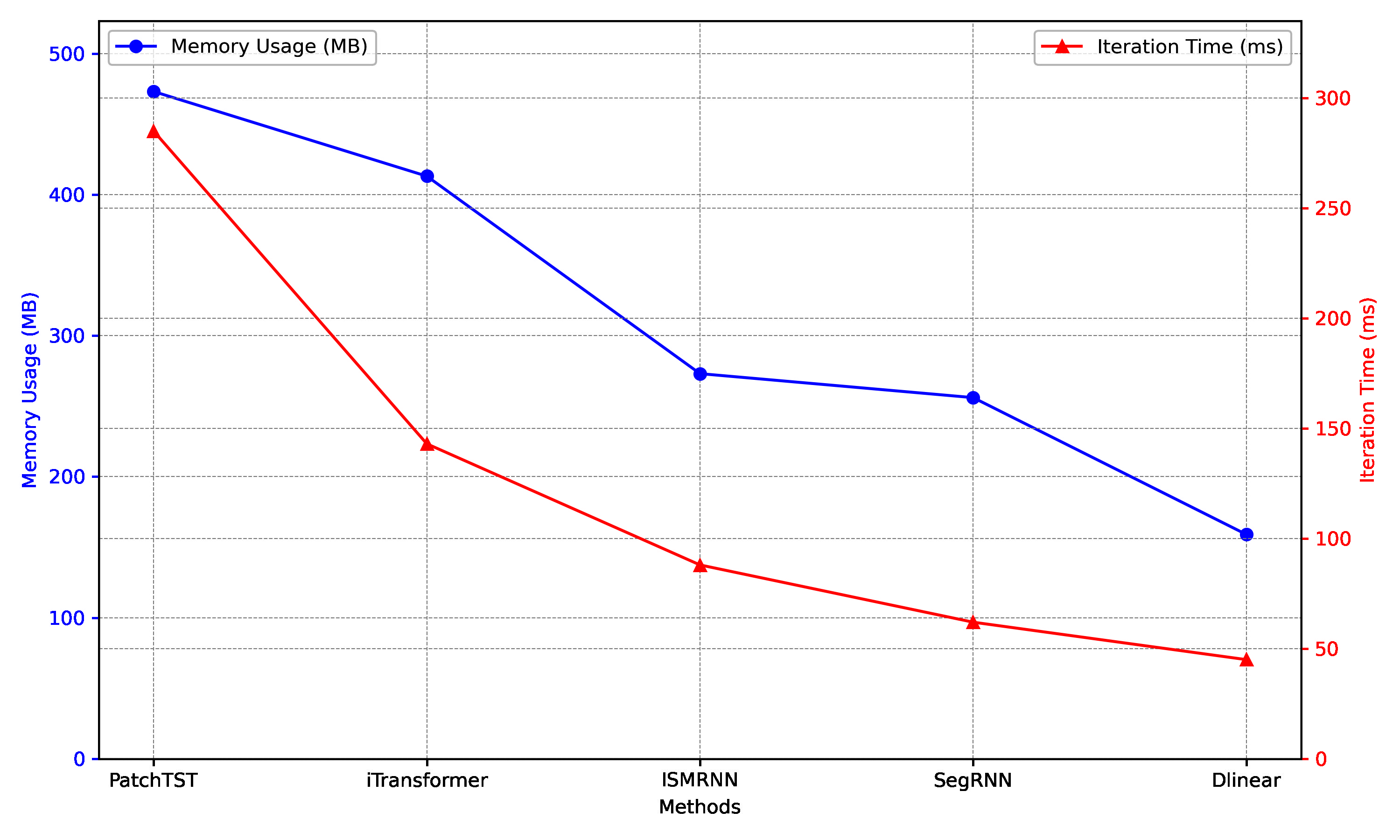}
	\caption{Comparison of spatiotemporal efficiency across different models.}
	\label{fig:time-memory-usage}
\end{figure}

This section is dedicated to evaluating the model's time and memory efficiency to substantiate its lightweight nature. The experiments are conducted on weather datasets because the other datasets utilize the Mamba architecture are implemented without convolutional layers and hardware optimization. The lightweight characteristic is primarily due to the use of a small number of parameters in our Mamba structure experiments. Using the Weather dataset as an example, the time and memory usage comparison with batch size 8 conducted on a single NVIDIA T4 GPU is shown in Figure \ref{fig:time-memory-usage}.

Compared to the SegRNN approach, our method introduces a slight increase in memory usage and training time, primarily due to the additional linear layers and the integration of the Mamba structure during the encoding phase. However, when compared to mainstream transformer-based models such as iTransformer and PatchTST, our model still demonstrates significant spatiotemporal advantages.

\section{Conclusion}\label{sec6}

In this study, we introduce the ISMRNN model, an implicitly segmented RNN method that integrates the Mamba structure, implicit segmentation, and residual structure. The Mamba structure aids the model in capturing long-term dependencies, addressing the gradient vanishing and long-term dependency issues inherent in RNNs. Implicit segmentation resolves the fixed segmentation problem of SegRNN by maintaining sequence continuity, while the residual structure in the model reduces information loss associated with the recurrent structure. These enhancements significantly improve performance in long-term time series forecasting while keeping memory usage and training time relatively low. Experiments on six real-world benchmark datasets validate our findings. Notably, the proposed implicit segmentation method is adaptable to various scenarios beyond the specific long-term time series forecasting problem. Future work will explore how these structures enhance model performance and identify the specific attributes of time series data where the ISMRNN method excels.

\bibliographystyle{unsrtnat}
\bibliography{main}  

\appendix
\section*{Appendix}

\section{Datasets}
For our Long-term Time Series Forecasting (LTSF) study, we employed several widely-used multivariate datasets, including:

ETTs: This collection consists of electricity transformer data from two counties in China, spanning two years. The datasets include various subsets: ETTh1 and ETTh2 provide hourly data, and ETTm1 and ETTm2 offer 15-minute interval data. Each entry encompasses the target variable "oil temperature" alongside six power load indicators.

Electricity: This dataset contains hourly electricity consumption records for 321 clients from 2012 to 2014, offering insights into consumer energy usage patterns over time.

Weather: This dataset captures 21 meteorological variables, such as temperature and humidity, at 10-minute intervals throughout 2020. It serves as a comprehensive resource for weather prediction and climate studies.

These datasets are instrumental for benchmarking and advancing forecasting models in the context of long-term time series analysis.

\section{Code Framework}

The implementation of our model framework is primarily derived from the official SegRNN code and the implementation of Mamba. This includes data preprocessing and evaluation, model parameter settings and device selection, as well as result output. We sincerely thank the authors of these open-source frameworks and methods for their contributions to our experiments. 

The official SegRNN source code is available at:
https://github.com/lss-1138/SegRNN.

The official minimal implementation of Mamba can be found at:https://github.com/johnma2006/mamba-minimal.

The official implementation of Mamba can be found at:https://github.com/state-spaces/mamba.git 

\section{Detail Configuration}

The main hyperparameter settings of our model are shown in Table \ref{tab:mseg_rnn_config}. The parameter \( d_{\text{state}} \) is a hyperparameter in the Mamba architecture, which is detailed in Mamba \citep{bib17}. The parameter settings and results for other methods are taken from their original papers or official code repositories.

\begin{table}[ht]
	\centering
	\caption{Detail configuration of ISMRNN}
	\fontsize{9}{11}\selectfont 
	\setlength{\tabcolsep}{1mm} %
	\begin{tabular}{lcccccc}
		\hline
		Datasets & d\_state & seg\_len & use\_conv & learning\_rate & dropout\\
		\hline
		ETTh1 & 2 & 12 & False & 0.0003 & 0.1\\
		ETTh2 & 4 & 24 & False & 0.0015 & 0.3  \\
		ETTm1 & 4 & 24 & False & 0.0010 & 0.1 \\
		ETTm2 & 4 & 24 & False & 0.0010 & 0.1 \\
		Weather & 4 & 24 & True & 0.0007 & 0.1 \\
		Electricity & 4 & 24 & False & 0.0014 & 0 \\
		\hline
	\end{tabular}
	\label{tab:mseg_rnn_config}
\end{table}

We also employed a different learning rate adjustment strategy. The learning rate remained constant for the first 15 epochs and then gradually multiplied by 0.9. This approach actually increases the learning rate compared to the adjustment method used in SegRNN.

\section{Comparison with extra methods}
We observed that some methods also utilized Mamba for information augmentation, we selected several long-term time series analysis models based on the Mamba architecture:
Time-Machine\citep{bib19}, Bi-Mamba4TS\citep{bib20}, and DT-Mamba\citep{bib21}, for comparison. The results are from official paper, respectivly . The result is shown in Table \ref{table:mamba_results}.

\begin{table*}[ht]
	\centering
	\caption{Performance of different models}
	\begin{tabular}{ccccccccccc}
		\hline
		Datasets & Models & \multicolumn{2}{c}{ISMRNN} & \multicolumn{2}{c}{TimeMachine} & \multicolumn{2}{c}{Bi-Mamba4TS} & \multicolumn{2}{c}{DTMamba} \\
		& & MSE & MAE & MSE & MAE & MSE & MAE & MSE & MAE \\
		\hline
		Weather & 96 & \textbf{0.147} & \textbf{0.187} & 0.164 & 0.208 & 0.164 & 0.212 & 0.171 & 0.218 \\
		& 192 & \textbf{0.198} & \textbf{0.238} & 0.211 & 0.250 & 0.214 & 0.256 & 0.220 & 0.257 \\
		& 336 & \textbf{0.256} & \textbf{0.282} & \textbf{0.256} & 0.290 & 0.269 & 0.296 & 0.274 & 0.296 \\
		& 720 & \textbf{0.334} & \textbf{0.337} & 0.342 & 0.343 & 0.348 & 0.349 & 0.349 & 0.346 \\
		\hline
		ETTh1 & 96 & 0.365 & \textbf{0.384} & \textbf{0.364} & 0.387 & 0.376 & 0.403 & 0.386 & 0.399 \\
		& 192 & 0.415 & \textbf{0.414} & 0.415 & 0.416 & \textbf{0.411} & 0.425 & 0.426 & 0.424 \\
		& 336 & 0.463 & 0.441 & \textbf{0.429} & \textbf{0.421} & 0.455 & 0.445 & 0.480 & 0.450 \\
		& 720 & 0.468 & 0.460 & \textbf{0.458} & \textbf{0.453} & 0.460 & 0.464 & 0.484 & 0.470 \\
		\hline
		ETTh2 & 96 & \textbf{0.275} & \textbf{0.326} & \textbf{0.275} & 0.334 & 0.289 & 0.341 & 0.290 & 0.340 \\
		& 192 & 0.354 & \textbf{0.378} & \textbf{0.349} & 0.381 & 0.367 & 0.389 & 0.366 & 0.392 \\
		& 336 & 0.404 & 0.416 & \textbf{0.340} & \textbf{0.381} & 0.410 & 0.424 & 0.380 & 0.409 \\
		& 720 & \textbf{0.407} & \textbf{0.428} & 0.411 & 0.433 & 0.421 & 0.439 & 0.416 & 0.437 \\
		\hline
		ETTm1 & 96 & 0.314 & \textbf{0.345} & 0.317 & 0.355 & \textbf{0.312} & 0.354 & 0.325 & 0.360 \\
		& 192 & 0.361 & \textbf{0.374} & \textbf{0.357} & 0.378 & 0.358 & 0.383 & 0.375 & 0.386 \\
		& 336 & 0.388 & \textbf{0.392} & \textbf{0.379} & 0.399 & 0.388 & 0.404 & 0.396 & 0.405 \\
		& 720 & 0.448 & \textbf{0.430} & 0.445 & 0.436 & \textbf{0.444} & 0.433 & 0.454 & 0.442 \\
		\hline
		ETTm2 & 96 & \textbf{0.170} & \textbf{0.248} & 0.175 & 0.256 & 0.174 & 0.259 & 0.177 & 0.259 \\
		& 192 & \textbf{0.233} & \textbf{0.292} & 0.239 & 0.299 & 0.240 & 0.304 & 0.240 & 0.300 \\
		& 336 & 0.294 & \textbf{0.331} & \textbf{0.287} & 0.332 & 0.303 & 0.345 & 0.310 & 0.345 \\
		& 720 & 0.391 & 0.389 & \textbf{0.371} & \textbf{0.385} & 0.402 & 0.407 & 0.395 & 0.394 \\
		\hline
		Electricity & 96 & 0.149 & 0.242 & \textbf{0.142} & \textbf{0.236} & \textbf{0.142} & 0.238 & 0.166 & 0.256 \\
		& 192 & 0.165 & 0.259 & 0.158 & \textbf{0.250} & \textbf{0.157} & 0.253 & 0.178 & 0.268 \\
		& 336 & 0.181 & 0.275 & \textbf{0.172} & \textbf{0.268} & \textbf{0.172} & 0.271 & 0.197 & 0.289 \\
		& 720 & 0.218 & 0.307 & 0.207 & 0.298 & \textbf{0.200} & \textbf{0.297} & 0.243 & 0.326 \\
		\hline
	\end{tabular}
	\label{table:mamba_results}
\end{table*}

It can be observed that our method performs notably well on the weather, ETTh2, and ETTm2 datasets, while the TimeMachine method shows significant advantages on the ETTh1 and Electricity datasets. Additionally, the Bi-Mamba4TS method also demonstrates a clear advantage on the Electricity dataset.

\section{Prediction Result}

We subsequently observed the partial prediction results of the model using the weather dataset, with a look-back window of 96 and a prediction horizon of 336. The results are shown in Figure \ref{fig:prediction_result}:

\begin{figure}[ht]
	\centering
	\includegraphics[width=0.23\textwidth]{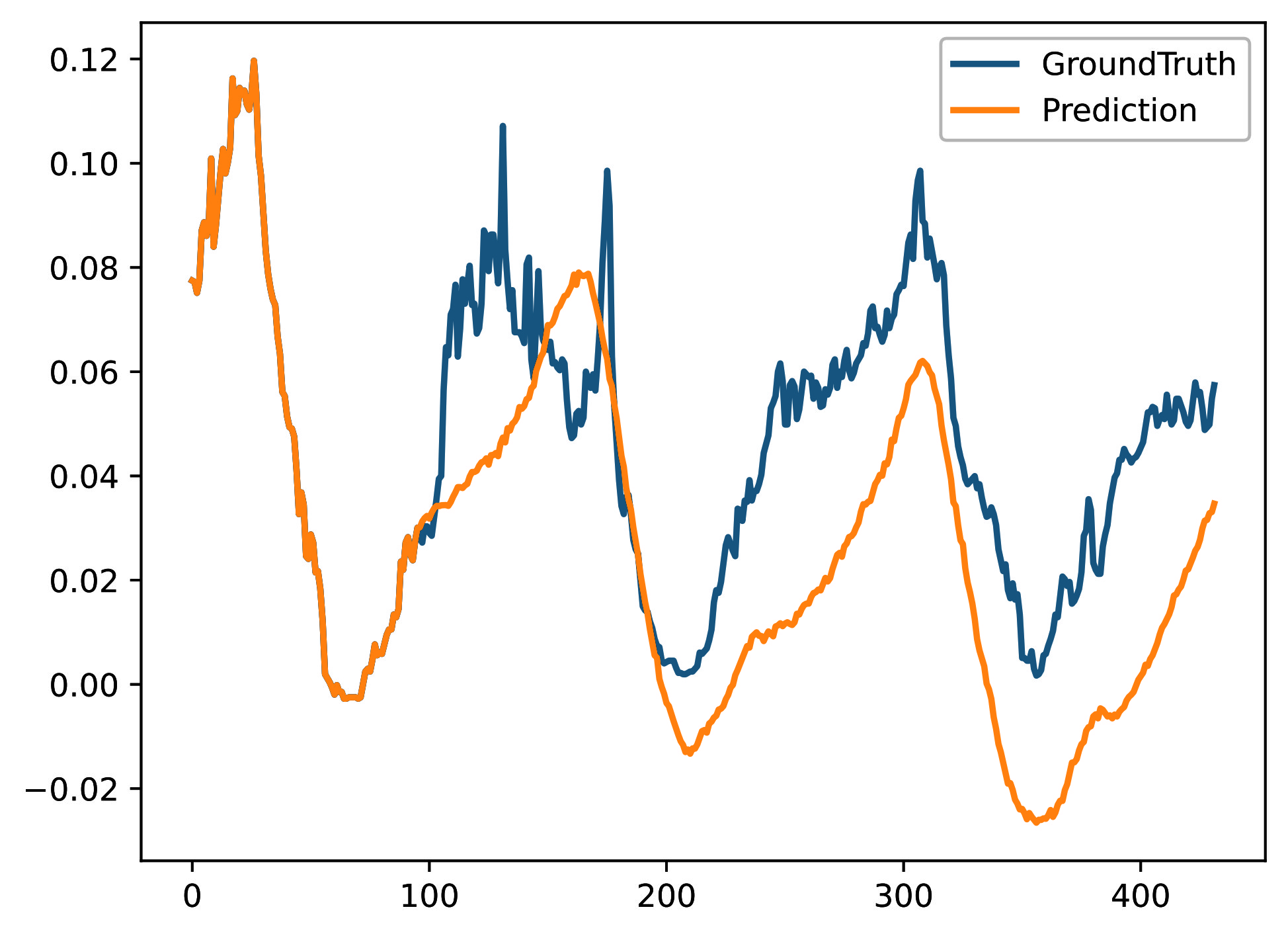}
	\includegraphics[width=0.23\textwidth]{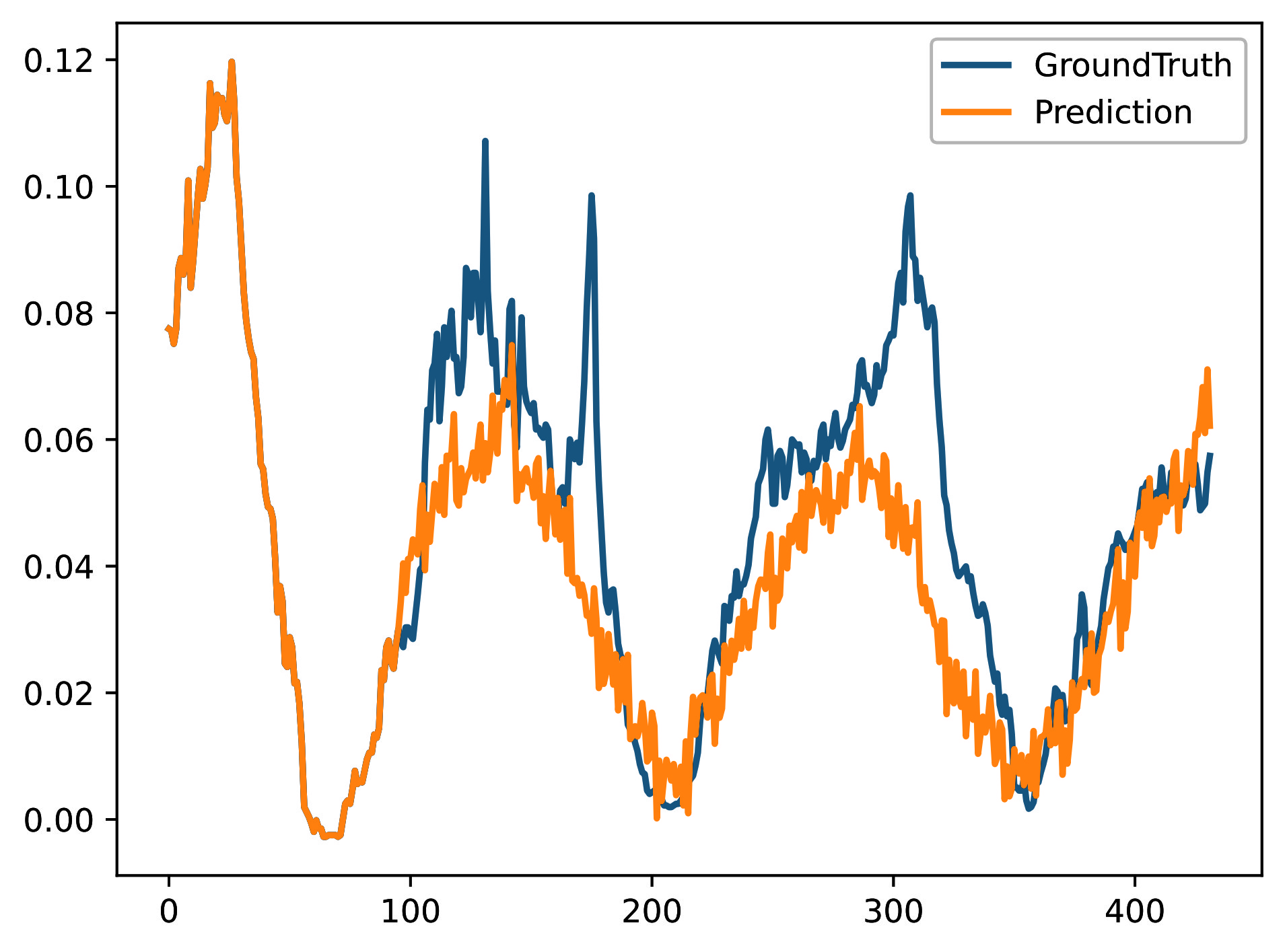}
	\includegraphics[width=0.23\textwidth]{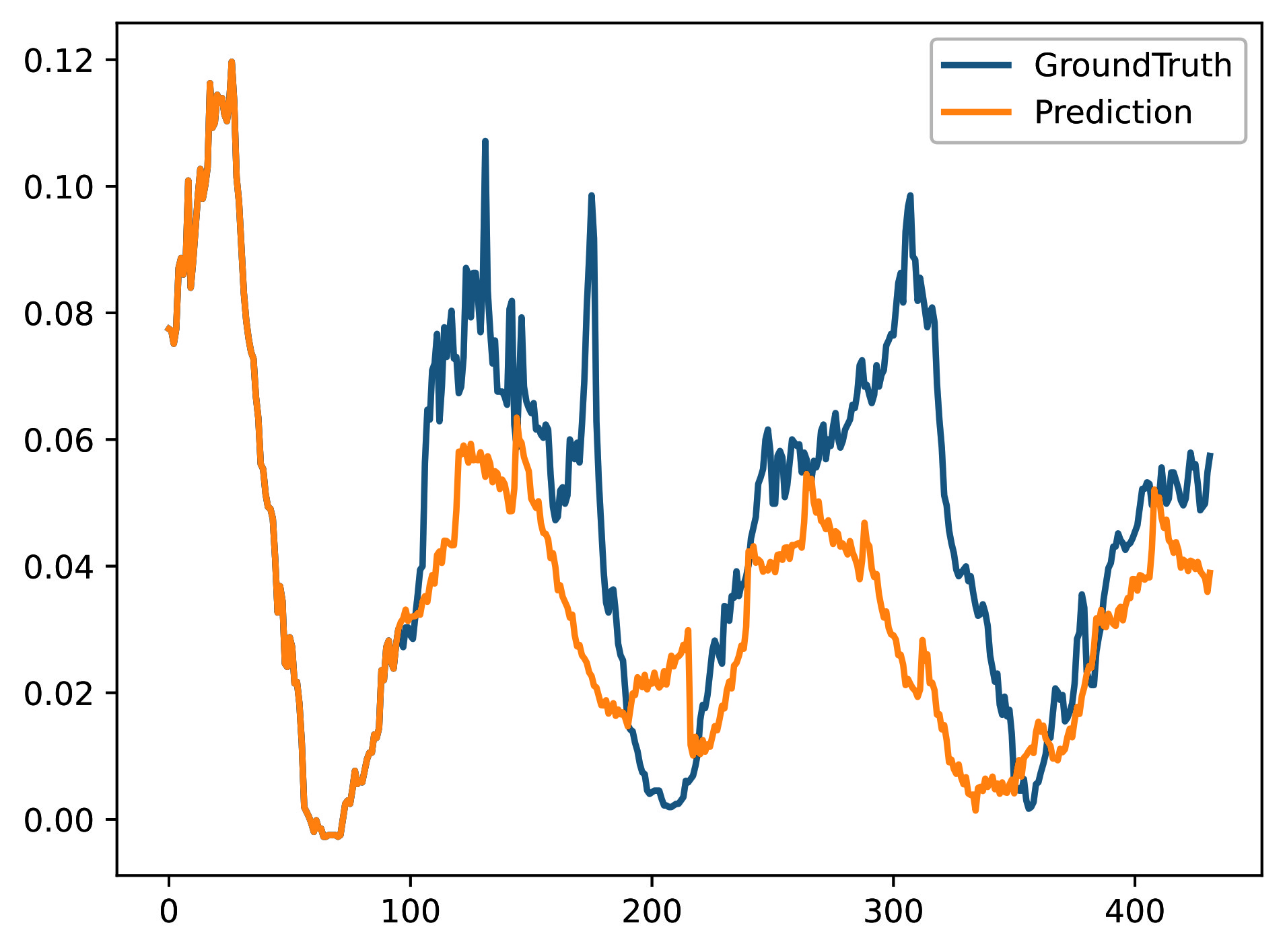}
	\includegraphics[width=0.23\textwidth]{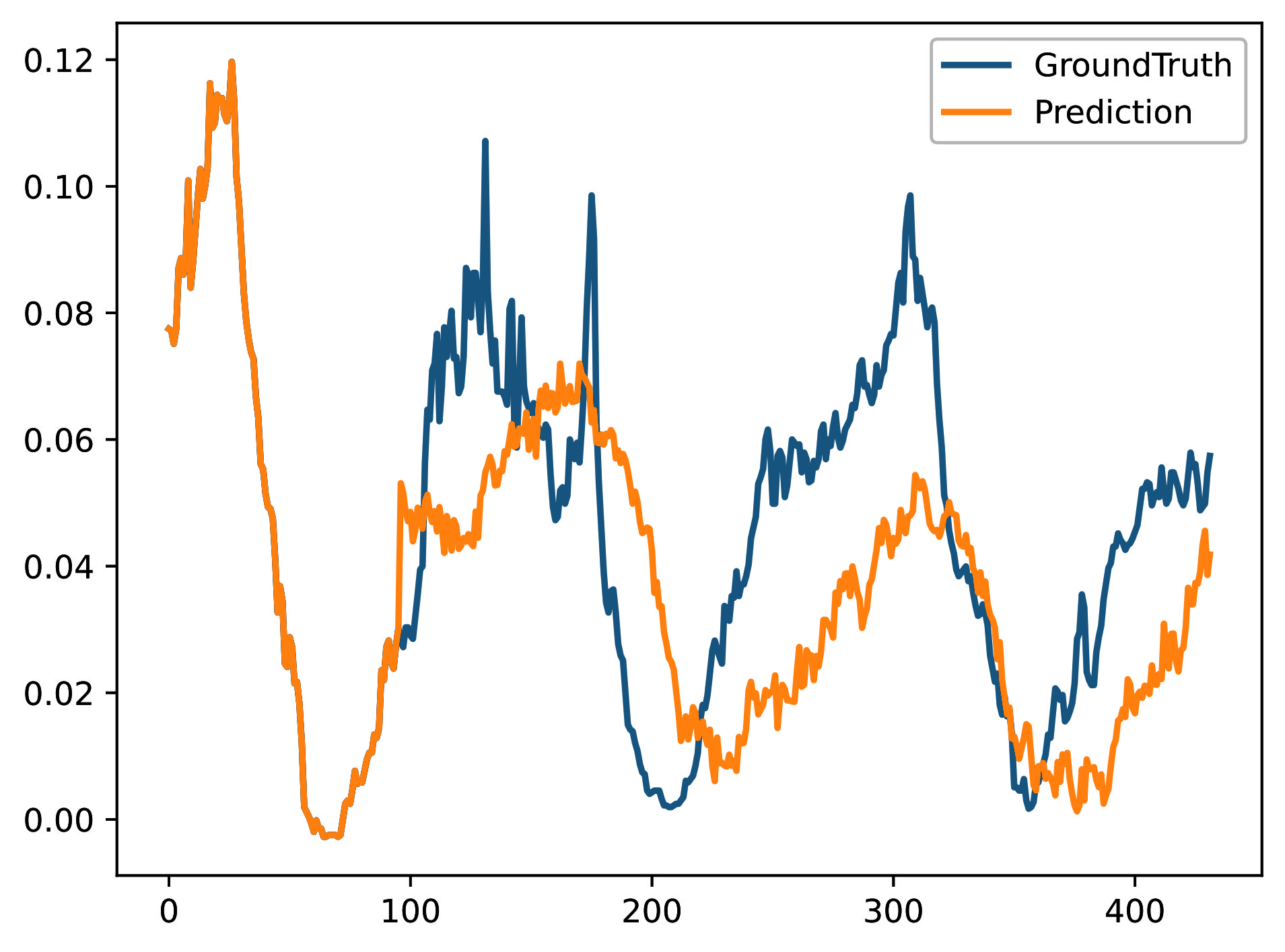}
	\caption{Prediction result of different method. The four methods are Dlinear, ISMRNN, SegRNN, and PatchTST in sequence. Intuitively, our method also has the best fitting performance, which is consistent with the fact that ISMRNN has the minimum loss among the models.}
	\label{fig:prediction_result}
\end{figure}

\section{Ablation study on using convolution layers}

In the main text's model structure, we removed the convolutional layer from the Mamba structure due to performance issues. We hypothesized that local features might have been redundantly captured. To investigate the impact of the convolutional layer on the model performence, we conducted experiments. The results are shown in Table \ref{table:conv_results}.

\begin{table}[ht]
	\centering
	\caption{Ablation study on convlution layer}
	\begin{tabular}{cccccccc}
		\hline
		Datasets & Models & \multicolumn{2}{c}{WithoutConv} & \multicolumn{2}{c}{WithConv} \\
		& & MSE & MAE & MSE & MAE \\
		\hline
		ETTh1 & 96 & \textbf{0.365} & \textbf{0.384} & 0.412 & 0.426 \\
		& 192 & \textbf{0.415} & \textbf{0.414} & 0.444 & 0.443 \\
		& 336 & \textbf{0.463} & \textbf{0.441} & 0.477 & 0.460 \\
		& 720 & \textbf{0.468} & \textbf{0.460} & 0.502 & 0.492 \\
		\hline
		ETTh2 & 96 & \textbf{0.275} & \textbf{0.326} & 0.282 & 0.338 \\
		& 192 & \textbf{0.354} & \textbf{0.378} & 0.372 & 0.396 \\
		& 336 & \textbf{0.404} & \textbf{0.416} & 0.406 & 0.424 \\
		& 720 & \textbf{0.407} & \textbf{0.428} & 0.430 & 0.456 \\
		\hline
		ETTm1 & 96 & \textbf{0.314} & \textbf{0.345} & 0.323 & 0.355 \\
		& 192 & \textbf{0.361} & \textbf{0.374} & 0.373 & 0.384 \\
		& 336 & \textbf{0.388} & \textbf{0.392} & 0.397 & 0.401 \\
		& 720 & \textbf{0.448} & \textbf{0.430} & 0.452 & 0.432 \\
		\hline
		ETTm2 & 96 & \textbf{0.170} & \textbf{0.248} & 0.180 & 0.264 \\
		& 192 & \textbf{0.233} & \textbf{0.292} & 0.240 & 0.305 \\
		& 336 & \textbf{0.294} & \textbf{0.331} & 0.299 & 0.343 \\
		& 720 & \textbf{0.391} & \textbf{0.389} & 0.396 & 0.399 \\
		\hline
	\end{tabular}
	\label{table:conv_results}
\end{table}

It can be observed that except for the Weather dataset in our experiments, the convolutional layer may interfere with the model's predictions in most scenarios, thereby limiting the model's capability.

\end{document}